\documentclass[11pt,a4paper]{article}
\usepackage[hyperref]{acl2020}
\usepackage{times}
\usepackage{latexsym}

\usepackage{microtype}
\usepackage{soul}
\usepackage[title]{appendix}

\aclfinalcopy % Uncomment this line for the final submission

\usepackage{float}
\restylefloat{table}

\title{What's in a Name?  Are BERT Named Entity Representations \\ just as Good for any other Name?}

\makeatletter
\renewcommand*{\@fnsymbol}[1]{\ensuremath{\ifcase#1\or \dagger\or \ddagger\or
    \mathsection\or \mathparagraph\or \|\or **\or \dagger\dagger
    \or \ddagger\ddagger \else\@ctrerr\fi}}
\makeatother

\author{Sriram Balasubramanian\thanks{\hspace{0.2cm}equal contribution, sorted alphabetically by last name} \hspace{.8cm} Naman Jain\footnotemark[1] \hspace{.8cm} Gaurav Jindal\footnotemark[1] \\
\textbf{Abhijeet Awasthi \hspace{.8cm} Sunita Sarawagi} \\
  Indian Institute of Technology, Bombay \\
  {\tt \small \{sriramb,namanjain,gauravj,awasthi,sunita\}@cse.iitb.ac.in}}
\usepackage{hyperref}
\usepackage{url}

\usepackage{microtype,defs}
\usepackage{graphicx}
\usepackage{subfigure}
\usepackage{booktabs} % for professional tables            
\usepackage{hyperref}

\usepackage{xcolor}

\usepackage{booktabs}
\usepackage{multirow}
\usepackage{siunitx}
\usepackage{tikz}
\usetikzlibrary{bayesnet}
\usepackage[ruled,vlined]{algorithm2e}

\usepackage[normalem]{ulem}

\usepackage{graphicx}
\RequirePackage[textsize=scriptsize]{todonotes}
\usepackage{mathrsfs, amsmath}
\usepackage{mathtools}

\newcommand{\green}[1]{\textcolor{blue}{#1}}
\newcommand{\red}[1]{\textcolor{red}{#1}}
\newcommand{\gray}[1]{\textcolor{gray}{#1}}

\begin{document}
\maketitle

\newcommand{\dict}{M}
\begin{abstract}
We evaluate named entity representations of BERT-based NLP models by investigating their robustness to replacements from the same typed class in the input. We highlight that on several tasks while such perturbations are natural, state of the art trained models are surprisingly brittle. The brittleness continues even with the recent entity-aware BERT models. We also try to discern the cause of this non-robustness, considering factors such as tokenization and frequency of occurrence. Then we provide a simple method that ensembles predictions from multiple replacements while jointly modeling the uncertainty of type annotations and label predictions. Experiments on three NLP tasks show that our method enhances robustness and increases accuracy on both natural and adversarial datasets.
\end{abstract}

\section{Introduction}
%\vspace{-0.3cm}
Contextual word embeddings from heavily pretrained language models \cite{peters2018deep, devlin2018bert} now form the basis of many NLP tasks.  While they have lead to improved accuracy for most tasks, %benchmarks have been broken on almost all fronts. %But while there have been advancements on many levels,
there are mounting concerns on how well these embeddings encapsulate syntactic and semantic constructs such as synonyms, misspellings, and knowledge representations. % and named entities. 
Indeed, it has been shown that even BERT based models are not robust to synonym swaps or spelling mistakes in a sentence \cite{jin2019bert, Lun2019onthe, sun2020advbert}. In this work, we investigate how well these contextual representations fare for named entities.

Designing robust representations of named entities 
% effectively has been a popular topic since long back \red{(??? need citation)}. 
is challenging due to the sheer variety of named entities. Named entities diversify with language, geographical location, time of history, and even with the fine types. Adding to this the varying length of such entities combined with out of vocabulary names, the complexity only increases. 

% \red{This paragraph is too much focused on attacks for representations?} Much recent work has studied the robustness of modern NLP models on natural perturbations of input sentences. Unlike for images, even defining the space of natural perturbations requires careful structuring and semantics. Thus, the space of perturbations has been limited to replacing words by their synonyms \cite{alzantot2018}, changing the surface form of verbs \cite{ribeiro2018}, and controlled paraphrasing \cite{iyyer2018paraphrase}.  The goal of these studies is to preserve the meaning of the sentence and keep it natural looking while breaking an existing pre-training model.

\begin{table*}[h!]
    \centering
    \begin{tabular}{|m{0.95\textwidth}|}
         \hline 
        %  Task: nlii \\
         \textbf{Sentence 1:} Magner , who is 54 and known as Marge , has been the consumer group 's chief operating officer since April 2002 , and sits on \sout{\green{Facebook}} \red{Microsoft} 's management committee \\
         \textbf{Sentence 2:} She has been the consumer unit 's chief operating officer since April 2002 , and sits \sout{\green{Facebook}} \red{Microsoft} 's management committee.
         \\
         \textbf{Gold:} 1 ;  \textbf{Prediction:} {\color{blue}Original}: 1; {\color{red}Perturbed}: 0\\
         {\centering \rule{{0.9\textwidth}}{0.5pt}}
         \\ 
         \textbf{Sentence 1:} The workers accuse \sout{\green{Goldman}} \red{Novell} of `` reverse age discrimination '' because of a change in retirement benefits in 1997 . 
         \\
         \textbf{Sentence 2:} \sout{\green{Goldman}} \red{Novell} was sued when it changed its retirement benefits in 1997 .
         \\
          \textbf{Gold:} 0 ;    \textbf{Prediction:} {\color{blue}Original}: 0; {\color{red}Perturbed}: 1
         \\
         \hline
    \end{tabular}
    \caption{Examples on paraphrase detection task -- Replacement of an entity }
    \label{fig:nli_ex}
\end{table*}

\begin{table*}[]
    \centering
    \renewcommand{\arraystretch}{1.25}%
    \begin{tabular}{|m{0.95\textwidth}|}
         \hline 
         \textbf{Task}: GEC; \textbf{Perturbed Entity}: Person \\
         \hline
        %  \textbf{Text:} \sout{\green{Michael}} \red{Hezekiah Wohlschlaeger} \textbf{get} away from there .\\
        %  \textbf{Original Prediction}: Michael \textbf{got} away from there .\\
        %  \textbf{Perturbed Prediction}: Hezekiah Wohlschlaeger \textbf{get} away from there .
           
        %  {\centering \rule{{0.9\textwidth}}{0.5pt}}
        %  \\ 
         \textbf{Text:}  One day \sout{\green{Penny}} \red{Bujalski} discovered it and it \textbf{go} to tell it to his queen . \\
         \textbf{Original Prediction}: One day Penny discovered it and \textbf{went} to tell it to his queen . \\
         \textbf{Perturbed Prediction}: One day Bujalski discovered it and go tell it to his queen.\\          
         {\centering \rule{{0.9\textwidth}}{0.5pt}}\\
         \textbf{Text:}  the two boys heard that he was planing to steal some money and kill people so the boys start their adventure on stopping \sout{\green{Abigale}} \red{Injuin Joe} . \\
         \textbf{Original Prediction}: The two boys heard that he was planning to steal some money and kill people so the boys started their adventure by stopping \textbf{Abigale} . \\
         \textbf{Perturbed Prediction}: The two boys heard that he was planning to steal some money and kill people so the boys started their adventure by stopping \textbf{Joe} . \\
         \hline
         \hline
         \textbf{Task}: GEC; \textbf{Perturbed Entity}: Country \\
         \hline
         \textbf{Text:} There are countries , such as \sout{\green{Greece}} \red{Oman} or \sout{\green{Bulgaria}} \red{Venezuela}  , in which the econmoy \textbf{relies} merely on tourism . \\
         \textbf{Original Prediction}: There are countries , such as Greece or Bulgaria , in which the econmoy \textbf{relies} merely on tourism .\\
         \textbf{Perturbed Prediction}: There are countries , such as Oman or Venezuela , in which the econmoy \textbf{rely} merely on tourism . \\
         {\centering \rule{{0.9\textwidth}}{0.5pt}}
         \\ 
         \textbf{Text:}   I am 20 years old \textbf{,} living in Port - Said , \sout{\green{Egypt}} \red{China} .  \\
         \textbf{Original Prediction}: I am 20 years old \textbf{and} living in Port - Said , Egypt . \\
         \textbf{Perturbed Prediction}: I am 20 years old \textbf{,} living in Port - Said , China .
         \\ 
         % If BERT remembers that Sardegna is in Italy then the edit below might even be correct.
        %  \textbf{Text:}  Last summer I went on holiday in Sardegna , \textbf{in} \sout{\green{Italy}} \red{Bhutan} , with my family .  \\
        %  \textbf{Original Prediction}: Last summer I went on holiday in Sardegna , Italy , with my family .  \\
        %  \textbf{Perturbed Prediction}: Last summer I went on holiday in Sardegna , \textbf{in} Bhutan , with my family.
        %  \\
         \hline \hline 
         \textbf{Task}: CoRef; \textbf{Perturbed Entity}: Person \\
         \hline
         \textbf{Text:} 
         And \green{\sout{Chris Hill}} \red{Sam Rusnock} our ambassador was in China a few days ago. he made the point and Secretary Rice made the point yesterday to the Chinese Foreign minister , we want to see China use its influence. Speaker Newt Gingrich the former speaker Republican weighed in on this debate in this way. [truncated] Well uh with all due respect to Speaker Gingrich we are on a course which has a reasonable chance of success.\\
         \textbf{Original Predicted Cluster}:  [``Chris Hill our ambassador",''he"] \\
         \textbf{Perturbed Predicted Cluster}: [``Sam Rusnock our ambassador",''he", ''Speaker Gingrich"] \\
         
        %  \textbf{Text:} If the President now changed his course won't the headlines be \green{\sout{Bush}} \red{Adams} blinks? No. I believe the headlines would be \green{\sout{Bush}} \red{Adams} [...truncated..].He reverses a policy that is not working.[...truncated..] uh I have nothing against the idea of the six party talks something Nick pounded.  \\
        %  \textbf{Original Predicted Cluster}:  [``the President",``Bush",``Bush", ''He"] \\
        %  \textbf{Perturbed Predicted Cluster}: [`'Adams",``Adams", ''He",''Nick"] \\
         
         {\centering \rule{{0.9\textwidth}}{0.5pt}}
         \\ 
         \textbf{Text:} 
         \green{\sout{Arianna Huffington}} \red{Sydnie Rabaut} uh in this lengthy piece this morning, Judy Miller is quoted excuse me as saying [truncated]. Do you buy this notion that she doesn't recall who this other source was? No of course not Howie. In fact I think this is the major unanswered question.\\                                     \textbf{Original Predicted Cluster}:[''Arianna Huffington", ''you", ''I"]\\
         \textbf{Perturbed Predicted Cluster}: [ ]
        %  And this Friday former president \green{\sout{Gerald Ford}} \red{Thomas Jefferson} will be ninety three years old. Our thirty-eighth president talked about age health and the presidency right here thirty one years ago. Mr. President in the early days [...truncated...]. \\
        %  \textbf{Original Predicted Cluster}:[''former president  Gerald Ford"], [''Our thirty-eighth president"], [''Mr. President"]\\
        %  \textbf{Perturbed Predicted Cluster}: [ ]
         \\
         \hline
    \end{tabular}
    \caption{Lack of robustness of GEC and CoRef model with respect to person and country names}
\label{fig:CoRefgec_ex}
\end{table*}
% \todo[inline]{CoRef: Syms example is all wrong. Syms is not a person name. I thought here we have gold annotations from Ontonotes.  Is Syms annotated as a person name?}

%For some category of NLP models, we want the prediction to not change even when the meaning of the sentence changes after perturbation. In this paper, we highlight one such important class of perturbations --- robustness to the substitution of name mentions in a sentence with other names within an entity class. 

We quantify how well current systems understand named entities by studying their robustness to substitutions of name mentions in a sentence with other names within an entity class. %adversarial examples for sentences and evaluate their performance.
The entity class within which we seek such robustness is task-dependent and easy for humans to provide. %as a form of high-level supervision. 
For example,  we may require a natural language inference model to be robust to the replacement of company names within the input sentence pairs. In Table~\ref{fig:nli_ex} we show a sentence pair which contains mentions of a company name {\tt Facebook}.  When we replace that mention with other company names like {\tt Microsoft} or {\tt Google}, a robust model should continue to make the same prediction. Likewise, we may require a co-reference resolution model to be robust to replacements of person names in a passage, and a grammar error correction model to be robust to replacement of person names of same gender or country names.  A good language representation should be able to generalize well to such perturbations and not deviate from its output upon such perturbations.% \todo[inline]{A good language representation should be able to generalize well to such perturbations ('not deviate from output' can be confusing }

%Previous work has evaluated NLP models to natural perturbations of input sentences. Unlike for images, even defining the space of natural perturbations requires careful structuring and semantics. Thus, the space of perturbations has been limited to synonyms swaps \cite{alzantot2018}, spelling errors \cite{hotflip}, changing the surface form of verbs \cite{ribeiro2018}, and controlled paraphrasing \cite{iyyer2018paraphrase}. In this work we highlight that named entity replacements while maintaining the structure of the sentence, can prove to be detrimental to model's predictions.

% In Section~\ref{sec:study}, 
The contributions of this work are three-fold. First, we investigate the robustness of trained NLP models using a generic algorithm that we develop.
%$G$ given an entity type class $c$ with a pool of members, and type annotations of input sentences. 
We empirically demonstrate a lack of robustness of state of the art BERT-based models
%that state of the art BERT-based models lack robustness 
for different user-specified typed classes spanning three NLP tasks: natural language inference (NLI), coreference resolution (CoRef), and Grammar Error Correction (GEC). The lack of robustness is specifically of concern for an entity-focused task like CoRef, where 85\% of test sentences have change in their predictions with a single person name change. At the same time we also found that Question Answering (SQuAD) was surprisingly completely robust to named entity attacks.

%\todo[inline]{corpus details can be defered}
%Our experiments were on the popular OntoNotes corpus, and our study calls for revisiting the way entity-focused NLP models get evaluated.

Second, we try to seek explanations for such lack of robustness, %testing frequency of occurrence of named-entities in the
by observing performance vs. frequency of named entities occurring in the fine-tuning dataset or based on the count of tokens in a named entity. We also explored if BERT's wordpiece-level masking was particularly unfavorable to entities by switching to Span-BERT, the recent span based masking model. While overall accuracy improved for all datasets with Span-BERT, we found no change in the robustness of the model.

Finally, we develop a simple approach that ensembles predictions from multiple replacements (RESEMBLE) while modeling the uncertainty of type annotations and label predictions.  Our approach not only improves performance on adversarial datasets but also on the original datasets, and achieves higher stability on all the tasks.

\section{Evaluating Robustness to Named-Entity Replacements} \label{sec:study}

% \red{While defining robustness and acceptable perturbation in the context of NLP models, previous work such as \cite{IBP} and \cite{Is Bert Really Robust?} have focused on ensuring that the perturbations are semantically and syntactically close to the original sentence. While this is a valid definition of robustness, we may want to expand the notion of “closeness” by using domain knowledge to define perturbations which should ideally not change the output of our NLP model. }

% \paragraph{Our Algorithm to Study Robustness}
%
% Let $V$ be a dictionary of candidate adversarial replacements. We measure the adversarial error of an example $\vx$ by perturbing it to a $\vx'$ that replaces a mention of the type in $\vx$ by the worst possible replacement from $V$.
% \todo[inline]{Reduce indent size. Algo looks too sparse now. What happened to budget?}
\begin{algorithm}[!t]
\SetAlgoNoLine
\SetArgSty{textnormal}
\newcommand\mycommfont[1]{\footnotesize\ttfamily\textcolor{blue}{#1}}
\SetCommentSty{mycommfont}
\KwData{$D$(dataset) , $V$(names), $M$(metric), $B$(budget)}
\KwResult{$D_{\text{worst}}$, $D_{\text{best}}$ (datasets on which the  \\ model performs worst and best)}
% $D_{\text{worst}} \longleftarrow \{ \}$ ;\ \  $D_{\text{best}} \longleftarrow \{ \}$ \;
\SetInd{0.1em}{0.5em}
 \For{$\left( x, y \right) \in D $}{
  min\_score $= \infty$,  max\_score $= -\infty$ ;\\
  $N \gets $ RandomSelection($V$, $B$) \;
  \For{$n \in N$}{
    $x' \gets $ Replace $\left( x , n \right)$ \tcp*{Details in text for each task}
    score $\gets M(G(x'), y)$ \;
    \If{$\text{score} < \text{min\_score}$}{
        min\_score $\gets$ score, 
        $~x_{\text{worst}} \gets x'$
    }
    \If{score $>$ max\_score}{
        max\_score $\gets$ score, $~x_{\text{best}} \gets x'$
    }
  }
  $D_{\text{worst}} \gets D_{\text{worst}} + \left( x_{\text{worst}}, y\right)$;\\  $D_{\text{best}} \gets D_{\text{best}} + \left( x_{\text{best}}, y\right)$\; 
 }
\caption{Probing a model using named-entity substitutions}
\label{algo:attack}
\end{algorithm}

We study the robustness of BERT-based NLP models w.r.t. type-specific named-entity substitutions,  for tasks like NLI, GEC, CoRef and SQuAD.  
%We study the robustness to named-entity substitutions over three NLP models built on BERT's contextual embeddings.  For each of the tasks, NLI, GEC and CoRef we study the robustness with to replacements from typed classes.
%We start by describing our algorithm for quantifying robustness.
Algorithm~\ref{algo:attack} describes our method of probing NLP models for lack of robustness. Let $V$ be a dictionary of candidate named entities of a given type $c$, and $D$ denote a dataset consisting of sentence-label pairs $\left(\vx, y \right)$. 
Let $G$ be a model fine-tuned on a pre-trained BERT.  For each sentence $(\vx,y) \in D$, we identify the mentions of named-entities of the type $c$ in $\vx$\footnote{We pre-filtered using a named entity tagger in the spaCy library, and made manual corrections so that all tagged entity mentions are correct in $D$.}. We obtain a perturbed sentence $\vx_m$ by replacing all mentions of a distinct name in $\vx$ by a random entry from $V$.  We repeat this process $B$ times where $B$ is a budget (we used 50), with replacement of names.  Over the $B$ perturbations, the sentence with the lowest accuracy is added to the set $D_\text{Worst}$ and the highest accuracy added to the set $D_\text{Best}$. A lower variance in model's performance across the datasets \{$D$, $D_\text{Worst}$, $D_\text{Best}$\} is indicative of higher robustness and vice-versa.  We also measure stability as the fraction of sentences in $D$ whose predictions stay unchanged within the budget sized replacements.

%Let $M(G(\vx), y)$ be the metric to measure the performance of the model. We first construct an adversarial dataset $D_{adv}$ by choosing the worst-performing name for each sentence out of sampling from $V$. Since the dictionary may be too large, we use random sampling to generate the adversarial sentences. We keep a maximum evaluation budget $N$, the maximum number of replacements done for a sentence. %The following algorithm assumes black-box access to the model.

We use the above method to evaluate the robustness of state-of-the-art BERT based models. We evaluate NLI with organization name replacements, GEC with person and country name replacements, and CoRef, SQuAD with person name replacements. 
In Table \ref{fig:bert_attack_result} we report accuracy on the original, worst, and best case perturbations of the input and stability for the four task-entity combinations. We discuss task details and results next.

%Now for each of our three tasks we describe the test set $D$, model $G$, and dictionary $V$ that were then processed by our algorithm in to generate the adversarial $D_\text{Worst}$ and favorable $D_\text{best}$ versions of original dataset $D$ along with the results.
\begin{table}[h]
\centering
\begin{tabular}{|c|c|c|c|c|}
\hline
 & NLI $\text{F}_1$ & \multicolumn{2}{c|}{GEC $\text{F}_{0.5}$} & CoRef $\text{F}_1$ \\ \hline
 Dataset & ORG & PER & COUN & PER \\ \hline
 Original & 84.82 & 50.93 & 47.87 & 76.47  \\ \hline
 Worst & 79.90 & 36.51 & 32.12 & 60.91 \\ \hline
 Best & 90.03 & 58.32 & 51.47 & 87.85  \\ \hline
 Stability & 86.8\% & 75\% & 63.4\% & 12.86\% \\ \hline
\end{tabular}
\caption{Adversarial Evaluation of BERT on different tasks}
\label{fig:bert_attack_result}
\end{table}

\subsection{Natural Language Inference (NLI)} 
\paragraph{Task} Paraphrase detection is a binary classification task on whether two sentences are paraphrases of each other. We work on the paraphrasing task of the GLUE dataset~\cite{glue2018}. The standard dataset split consists of 4077 training sentence pairs and 1726 testing pairs. We use the BERT-base model fine-tuned on the training dataset. The model takes as input the concatenated sentence pairs and predicts a binary output. The metric used for this task is $\text{F}_1$ score on the binary output. 

\paragraph{Attack details} We measure robustness over the organization concept class. As the replacement dictionary $V$ we used organization names from Fortune 500 companies.  We filter out sentence pairs consisting of organization name mention in each sentence of the pair and get 218 sentence pairs. We use spaCy~\cite{spacy} for tagging the sentences followed by manual inspection of matched entities so that in the 218 filtered sentences all entity mentions are correctly identified.  

\paragraph{Results} Observe in Table~\ref{fig:bert_attack_result} almost a 10\% swing in F-score between $D_\text{Worst}$, $D_\text{Best}$  just by replacing organization names in test instances. The perturbation dictionary consisted of Fortune 500 companies, and were not particularly obscure either. As the examples in Table \ref{fig:nli_ex} show some of these replacements do not span rare names (\texttt{Facebook} to \texttt{Microsoft} or \texttt{Goldman} to \texttt{Novell})

\subsection{Grammatical error correction (GEC)} 
\paragraph{Task} Grammatical error correction is a sequence prediction task, given an incorrect sentence as input we have to predict the grammatically correct output. We use the LOCNESS corpus \cite{locness} comprising of incorrect and correct parallel English essays. The standard dataset split consists of 34,308 incorrect-correct sentence pairs for training and 4,384 pairs for testing. We use two types of GEC models to analyze the performance of GEC. Our first model is the parallel edit model from \cite{awasthi2019parallel} which uses a BERT model for predicting the edits at every token on the input and applies those edits to compute the final output. We only use a single iteration of the model for ease of evaluation. Our second model is a sequence to sequence prediction model from \cite{zhao2018generating} which aligns predictions and input by learning explicit copy scores. The performance is measured using $\text{F}_{0.5}$ score based on M2 files \cite{errant}.

\paragraph{Attack details} For the BERT based GEC model we measure robustness on two concept classes: person names and country names. %\todo{EMNLP: Gender of names? Done?} 
%Using spacy, we filter out sentences containing 2 typed classes -- person and countries followed by coarse manual filtering.
From the test set, 328 sentences mentioned person names and 82 mentioned country names. For person names, we perform gender-specific replacements. The person name dictionary was created as follows:  we start with a large dictionary of 4018 female first names, 3437 male first names and 151670 last names and remove names encountered in the training data. We then generate about 250 names from these sets by combining first names and last names. For countries we use 58 non-frequent country names. For Seq2Seq GEC model performance is evaluated on person name class.  %\todo{EMNLP: Why only non-frequent, best might improve if frequent also allowed}  

\paragraph{Results} For BERT-GEC model as we can see from \ref{fig:bert_attack_result} the gap in accuracy between the best and worst-case perturbations is almost 20\% for both person name and country name replacements. Moreover, we find that 25\% of the sentences change prediction on changing person names and more than 35\% sentences vary prediction of country names! Even the Seq2Seq GEC model exhibits similiar non robustness -- on $D_{worst}$ the F score dropped from 42.3 to 36.7 while on $D_{best}$ it increased upto 45.24. Table \ref{fig:CoRefgec_ex} shows some examples of BERT-GEC model.  Notice how changing the country from {\tt Greece} to {\tt Oman} and {\tt Bulgaria} to {\tt Venezuela} changes the edit predictions five tokens away in the sentence. 

\subsection{Coreference Resolution (CoRef)} 
\paragraph{Task} Coreference resolution refers to the problem of finding all expressions that refer to the same entity in a text. We work on the standard OntoNotes dataset from the CoNLL-2012 shared task on coreference resolution \cite{pradhan2012CoRef}. Each document represents one  instance and has a series of sentences within it. The standard split consists of 2,802 training documents and 348 testing documents. We use the BERT base model fine-tuned on the training dataset from \cite{joshi-etal-2019-bert}. The model predicts top-k spans for a document and then computes antecedent scores for them and thereby builds clusters for coreference. Since documents in OntoNotes contain many clusters while we replace only mentions of a single name in the long document, to better highlight differences, we measure F score for only the gold clusters with the replaced entity. 

\paragraph{Attack details} We measure the robustness with respect to person names. We filter out documents containing a person name based on gold annotations in the OntoNotes corpus, and get 210 documents. Replacement vocabulary $V$ was made in similar way as mentioned for GEC using the same male, female and last name dictionaries. %\todo{EMNLP: Might have been nice to use the same person name dictionary for all tasks.  Also, the perturbation details have to be put along with the algorithm description and not repeated here.}   
% \todo[inline]{Did we do manual inspection to remove sentences encode world-knowledge about a person, e.g., president? --> Yes}
We also ensure that the name replacements do not alter the coreferences. Therefore, we replace every instance of each name occurring in the document with our randomly sampled adversarial name, taking care that first(or last) names are replaced with adversarial first(or last) names. In case of any ambiguity, we replace the name with the last name. Also the replacements are gender specific.

\paragraph{Results} We found the worst stability for CoRef and only 13\% of the sentences preserved predictions on named-entity replacements.  Also, the gap between the worst and best case perturbations is almost 30 $\text{F}_1$ points. As seen from the truncated document examples in the second-last row of Table \ref{fig:CoRefgec_ex}, replacing the name {\tt Chris Hill} to \texttt{Sam Rusnock} makes the model mis-predict the original cluster, as it predicts another name {\tt Speaker Gingrich} as co-referent to {\tt Sam Rusnock}. Even in second example changing the name {\tt Arianna Huffington} to {\tt Sydnie Rabau} causes model to miss the its entire cluster! We also found that on an average, predictions of model differ by two clusters per sentence after name perturbation. For one document almost 17 clusters were affected by a single entity swap.
The non-robustness on CoRef is especially surprising since it is principally a task about named entities. Our experiments were on the widely used OntoNotes dataset with person name mentions.  Such varying performance should be a cause of concern for benchmarking CoRef models. Perhaps, the dataset needs to be augmented with variants arising out of named-entity replacements and stability should be a required performance metric, in addition to accuracy on the original sentence. 

%As we can see from the results above all three tasks there is a deviation in performance of models with named entity replacements. The performance drop is more drastic for structure prediction tasks (CoRef and GEC) compared to binary prediction tasks like NLI that present a simpler decision surface learnable by other words in a sentence. 

Another interesting observation from these results is that across tasks is that the accuracy on the original $D$ is enhanced after moving to $D_\text{Best}$ --- that is, just substituting names in a given instance with more `favorable' names can lead to substantial gains. We will exploit this observation to enhance base accuracy and improve the robustness of NLP models in Section~\ref{sec:enhance}.

\subsection{SQuAD }
\paragraph{Task} Squad/Question answering refers to the problem of finding the correct answer to a question give a reference text. We work on the standard Squad 1.0 dataset \cite{rajpurkar2016squad}. The dataset consists of articles/documents containing multiple paragraphs and multiple questions per each paragraph. The standard split consists of 550 training articles and 55 testing articles. We use the BERT base model fine-tuned on the training dataset from \cite{devlin2018bert}. Performance is measured using F-score.

\paragraph{Attack details} We measure the robustness with respect to person names. We filter out paragraph question pair such that each question contains a person name. We appropriately replace all presence of a name in paragraph (to avoid altering coreferences) while taking care of gender of replacement. We used a budget of 10.  %\todo{EMNLP: Might have been nice to use the same person name dictionary for all tasks.  Also, the perturbation details have to be put along with the algorithm description and not repeated here.}   

\paragraph{Results} Contrary to tasks such as NLI, GEC, and CoRef, we found SQuAD to be robust to named entity perturbations. On the restricted dataset containing person name in the question, BERT had an F-score of 82.0. After performing the replacements defined above the F-score only dropped to 81.9.

% \begin{table}[h]
% \centering
% \begin{tabular}{|c|c|c|c|c|c|}
% \hline
% & Original & Worst & Best & Stability \\ \hline
% GEC(Person)  &     50.59         & 38.48   & 57.08 & 75\%    \\ \hline
% GEC(Country) &     47.87          & 32.12 & 51.47 & 63.4\% \\ \hline
% \end{tabular}
% \caption{Adversarial evaluation of bert based gec model}
% \label{fig:bert_gec_result}
% \end{table}

% \begin{table}[h]
% \centering
% \begin{tabular}{|c|c|c|c|c|c|}
% \hline
% & Original & Worst & Best & Stability  \\ \hline
% Person  &        78.28        &             59.85 & 87.57 & 20.24\%   \\ 
% \hline

% % & Original Test & Adversarial Test & Stability(\%) \\ \hline
% % Person      &        80.35        &             77.16      &        29.63   \\ \hline
% \end{tabular}
% \caption{Adversarial evaluation of bert based CoRef model}
% \label{fig:bert_CoRef_result}
% \end{table}

%in natural language inference (NLI), grammatical error correction (GEC) and CoReference resolution (CoRef). 
% \todo[inline]{How is the adversarial dictionary obtained? Put in details here of discounting entities that are frequenting in training data etc.}
%\todo[inline]{--> add N is budget}

% \begin{table}[!h]
% \centering
% \begin{tabular}{|c|c|c|c|c|}
% \hline
%  & Original & Worst & Best   & Stability  \\ \hline
% Organization  &     85.18         & 81.18   & 89.13 & 90.3\% \\ \hline
% \end{tabular}
% \caption{Adversarial evaluation of bert based NLI model}
% \label{fig:bert_nli_result}
% \end{table}

\begin{figure}
\includegraphics[width=\linewidth]{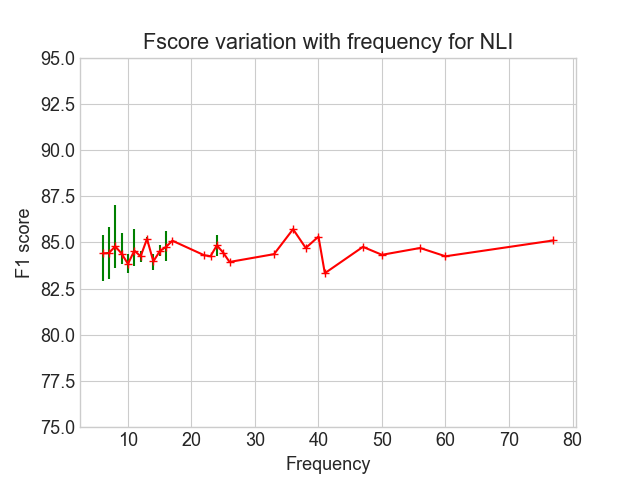}
\caption{Variation of NLI model's performance with frequency of named entity in training dataset. The green lines depict variance across performance of names of a given frequency}
\label{freq}
\end{figure}

\section{Causes of Non-Robustness}
We then sought to investigate reasons for such lack of stability. 
We first attempted to see if the poor accuracy of certain names can be explained by their frequency of occurrence in the training dataset.  In Figure~\ref{freq} we plot a graph of the frequency of a named-entity in the training corpus against the F-score on the NLI task. As we can see there is no strong correlation of frequency with the performance of a named entity, in fact, an organization name appearing in only four sentence pairs (\texttt{Goldman}) performed better than \texttt{Microsoft} which was present in over 30 sentence pairs. \texttt{Facebook} which is not even present in the training set performs better than \texttt{Microsoft} or \texttt{Google}. This is likely due to the biases learned during the massive pre-training that BERT-based models enjoy.

\begin{figure}
\includegraphics[width=\linewidth]{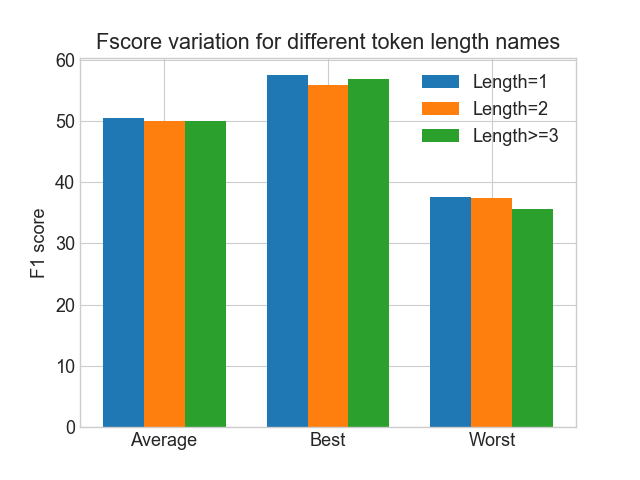}
\caption{Variation of $\text{F}_1$ scores for GEC model with different token length perturbation.}
\label{tokenlen}
\end{figure}

Our next guess was to see if the number of tokens in BERT's word-piece tokenization of named entity causes any significant impact on accuracy.  Sequence labeling models like PIE \cite{awasthi2019parallel} for GEC are most likely to be susceptible to that effect. In Figure~\ref{tokenlen} we show accuracy against the number of tokens in a named entity for GEC. We compared performance across three classes -- 1 token length entities or two token length entities or three or more token length entities. We created budget sized copies of the original dataset and compare performance across three variants -- (Original, Best, and Worst) but found no significant difference in accuracy with the number of tokens. However, we did observe some anecdotal evidence of specific nuisance tokens arising out of the word piece model on out of vocabulary names. For example consider the person name {\tt Tobey} that gets tokenized as {\tt [To, \#\#bey]} or {\tt Injuin} which is tokenized as {\tt [In, \#\#juin]}. The first token of the names are ``{\tt To}'' or ``{\tt In}'', both frequent prepositions,  which perhaps BERT finds difficult to disambiguate. As we can see from the second example in Table \ref{fig:CoRefgec_ex} -- {\tt Injuin} confuses the given model and the model even deletes the name probably since ``In'' proposition is not required there. Another artifact could be memorized correlations between names (e.g. {\tt Obama} and {\tt President}) that tasks like CoRef could exploit. Recent work \cite{poerner2019} has infact shown that BERT based models use surface form of entities for relational reasoning.

%We conjecture that this effect is caused by training artifacts arising during training and finetuning of bert. The training datasets \red{along with randomness arising in during training (through initialization, data permutation, dropouts)} introduce biases in the model.\\
%Consider the two named entities -- \texttt{James} and \texttt{Obama}. While both of the names correspond to a person, during training bert, \texttt{Obama} would mostly come in political context which would be significantly different from \texttt{James}. Hence expecting such robustness from bert based models is over-expectation. 

\begin{table}[h]
\centering
\begin{tabular}{|c|c|c|}
\hline
 & NLI (ORG) $\text{F}_1$ & CoRef (PER) $\text{F}_1$\\ \hline
 Original & 86.80 & 76.71  \\ \hline
 Worst & 82.7 & 62.37 \\ \hline
 Best & 90.2 & 86.76  \\ \hline
 Stability & 89.9\% & 16.19\% \\ \hline
\end{tabular}
\caption{Adversarial Evaluation of Span-BERT on different tasks}
\label{span_attack_result}
\end{table}

Finally, we explore if BERT's single token masking model is unfavorable to robust entity representations by comparing with a language model pre-trained by masking spans covering multiple tokens. Specifically, we use Span-BERT \cite{joshi2019spanbert}, which is trained with masked language modeling on spans instead of tokens. We tried to compare the performance on NLI and CoRef\footnote{We were unable to train Span-BERT for GEC, since in released Span-BERT checkpoints were not compatible with the GEC model} in comparison with BERT. % \red{We also consider the named entity and noun variant of the model where 50\% of spans are named entities and nouns respectively which could be considered more entity aware of their training procedure.} 
The results can be found in Table \ref{span_attack_result}.
We were surprised that Span-BERT does not provide any better robustness, although it does provide consistent higher accuracy on all tasks. Various metrics such as -- the difference between worst and best accuracy, stability are both very similar for BERT and Span-BERT. 
% For CoRef, the stability for Span-BERT model is infact lower (17.06 \%) compared to BERT (20.24 \%)

%\begin{table}[!h]
%\centering
%\begin{tabular}{|c|c|c|c|c|}
%\hline
% & Original & Worst & Best & Stability \\ \hline
%F score      &     88.0         & 82.75   & 90.20 & 89.90\% \\ \hline
%\end{tabular}
%\caption{SPAN BERT NLI RESULTS}
%\label{fig:span_nli_result}
%\end{table}
%
%\begin{table}[!h]
%\centering
%\begin{tabular}{|c|c|c|c|c|}
%\hline
% & Original & Worst & Best & Stability \\ \hline
%F score      &     79.2         & 60.23   & 88.62 & 17.06\% \\ \hline
%\end{tabular}
%\caption{SPAN BERT CoRef RESULTS}
%\label{fig:span_CoRef_result}
%\end{table}

%\subsection{Robustness via replacements}
\vspace{-2pt}
\section{Enhancing Robustness} \label{sec:enhance}
\vspace{-2pt}
We propose a simple ensembling with replacements approach (referred to as RESEMBLE) that does not require any retraining and can work with any existing pre-trained language model. We assume a type annotator $T$ that marks mentions of entities of the type $c$ for which robustness needs to enhanced. The type-annotator might be noisy. We identify a small set $M$ of entities of type $c$ on which the model provides high accuracy on a validation set.  We call these the list of canonical entities.  

%We detect named entities in a sentence and replace the named entities (identified by a named entity tagger) with a canonical named entity commonly found in the training dataset. For example, we might replace all person names with a common name like John or Mary. This defense strategy works quite well in practice (see Col. 1 in \ref{defense tables}). Almost all of the performance loss from name substitution attacks can be recovered and sometimes the nominal performance even improves on using the canonical name.

%The above approach assumes access to a perfect named entity tagger which is not true. It can be improved by using the confidence of named entity tagger and model output in a graphical model. 

Given any input $\vx$, we invoke the task-specific model $G$ to obtain predicted labels $\hat{\vy}$ and the type annotator $T$ to obtain type annotations $\hat{\vz}$. If $\hat{\vz}$ denotes that a named entity of type $c$ is present in one or more spans of $\vx$, we generate new sentences $\vx_m$ by replacing the named entities with canonical named entities $m \in M$. The model $G$ when applied to $\vx_m$ generates prediction $\hat{\vy}_m$. % and types $\hat{\vz}_m$ respectively.  

Let the true labels of $\vx$ and $\vx_m$ be $\vy$ and $\vy_m$ respectively, and the true type of $\vx$ be $\vz$. % and $\vz_m$. 
If the type annotator correctly identified the spans corresponding to concept class $c$ (i.e., $\vz=\hat{\vz})$,  $\vy$ and all $\vy_m$s have to agree as per our requirement of robustness. We use this to define a revised distribution over true $\vy$ from the individual predictions as follows:

\begin{equation}
\begin{split}
    P_R(\vy|\vx,\hat{\vz}) \propto (1-P(\vz=\hat{\vz}|\vx))P(\vy|\vx)\\ + P(\vz=\hat{\vz}|\vx) \left(P(\vy|\vx)\prod_m P(\vy|\vx_m)\right)^{\frac{1}{m+1}}
\end{split}    
\end{equation}
The above is an annotator confidence weighted average of two terms:
The first half calculates the probability of $\vy$ from the default model $G$ when the type annotator may be wrong and the $\vy_m$ predictions should be ignored.  The second half calculates the ensembled agreement probability when the type annotator is correct.  We calculate that as a geometric mean of the predictions from the different replacements. In the above equation, the ensembled probability is under the simplifying assumption that all entity replacements have the same number of tokens.  During implementation, we remove this assumption, and implement a more detailed span-level agreement for variable-length entities.%\todo{Might be useful to give the exact details in the Appendix.}

%Intuitively, we need to maximize both robustness and nominal accuracy. The arg max of the agreement probability is robust to name replacements, while the arg max probability from the base model is more likely to be accurate. Because the type-annotator may not be perfectly accurate, we need to weigh both the terms by the confidence of the type tagger. Instead of naive multiplication, we replace it with the geometric mean of the predictions to accommodate the diminishing magnitude of the product. 

An important requirement for the above expression is that the probabilities provided by the different models express true uncertainty of predictions, that is, they be well-calibrated.  Unfortunately, modern neural networks tend to be uncalibrated. To calibrate the probabilities, we use a popular method called temperature scaling \cite{GuoPSW17} where probabilities are raised by an exponent, which is the inverse of the temperature. Temperature scaling flattens the probability distribution over output classes thus reduces the confidence until it is correctly calibrated. The expression is as follows:
$$P_T(y|\vx)=\frac{P(y|\vx)^{\frac{1}{T}}}{\sum_{y'}P(y'|\vx)^{\frac{1}{T}}}$$
where $y$ denotes a scalar prediction.  For two of our tasks (GEC and CoRef), the output from our BERT-based models is a product of probabilities from multiple positions.  We apply the same temperature scale to each prediction. Thus, our final expression becomes:

%and multiply a constant scale (S) to the first term in the above expression. So we use a validation dataset to both (a) choose $M$ and (b) learn a joint calibration of the different distributions. The final expression becomes

\begin{equation}
\begin{split}
    P_R(\vy|\vx,\hat{\vz}) \propto (1-P(\vz=\hat{\vz}|\vx))P_T(\vy|\vx)\\ + P(\vz=\hat{\vz}|\vx) \left(P_T(\vy|\vx)\prod_m P(\vy|\vx_m)\right)^{\frac{1}{m+1}}
\end{split}    
\end{equation}

% \begin{equation}
% \begin{split}
%     P_R(\vy|\vx,\hat{\vz}) \propto P_T(\vy|\vx)\Big((1-P_{T_z}(\vz=\hat{\vz}|\vx))\\ + P_{T_z}(\vz=\hat{\vz}|\vx) \prod_m P(\vy|\vx_m)^{\frac{1}{m}}\Big)
% \end{split}    
% \end{equation}
The temperature hyper-parameter $T$ is fixed from a validation dataset.  Note we do not apply temperature scaling to the predictions from the canonical entries. %because these are the 'few chosen high-quality' replacements on which over-confidence did not appear to be an issue.

% \todo[inline]{Write the exact formula for recalibration here. If we recalibrate using the validation dataset so that less accurate $m$ have a flatter distribution, I think the above method will work quite well.}

\subsection{Empirical Results}

\begin{table*}
\centering
\begin{tabular}{|c|c|c|c|c|c|c|c|c|c|}
\hline
Dataset & \multicolumn{3}{c|}{NLI(ORG) $\text{F}_1$} & \multicolumn{3}{c|}{GEC(PER) $\text{F}_{0.5}$}  & \multicolumn{3}{c|}{CoRef(PER) $\text{F}_1$} \\ \hline
% & \multicolumn{2}{c|}{ORG} & \multicolumn{2}{c|}{PER}  & \multicolumn{2}{c|}{PER} \\ \hline
 & \multirow{2}{*}{Original} &  \multicolumn{2}{c|}{RESEMBLE} & \multirow{2}{*}{Original} & \multicolumn{2}{c|}{RESEMBLE} & \multirow{2}{*}{Original} & \multicolumn{2}{c|}{RESEMBLE} \\ \cline{3-4} \cline{6-7} \cline{9-10}
& & M=1 & M=3 & & M=1 & M=3 & & M=1 & M=3 \\ \hline
Original & 84.80 & 85.16 & 85.16 & 50.93 & 51.81 & 51.53 & 76.47 & 76.87 & 76.71 \\ \hline
Worst & 79.90 & 82.71 & 82.71 & 36.51 & 47.09 & 46.75 & 60.91 & 68.31  & 69.43 \\ \hline
Best & 90.03 & 86.91 & 86.63 & 58.32 & 55.38  & 55.38 & 87.85 &  82.6 & 82.18 \\ \hline
Random & 85.53 & 85.48 & 85.50 & 49.96 & 51.47 & 52.03 & 76.37 & 76.78 & 76.87 \\ 
Replacement & \small \gray{(0.54)} & \small \gray{(0.50)} & \small \gray{(0.40)} & \small \gray{(1.05)} & \small \gray{(0.70)} & \small \gray{(0.66)} & \small \gray{(1.04)} & \small \gray{(0.75)} & \small \gray{(0.81)} \\\hline
% Original & 84.80 & 50.59 & 47.87 & 78.28  \\ \hline
% Worst & 79.9 & 38.48 & 32.12 & 59.85 \\ \hline
% Best & 90.03 & 57.08 & 51.47 & 87.57  \\ \hline
% Stability & 86.8\% & 75\% & 63.4\% & 20.24\% \\ \hline
\end{tabular}
\caption{Adversarial Evaluation of BERT on different tasks comparing the accuracy on the original model against our algorithm with a canonical dictionary of size ($M$) 1 or 3. For Random Replacement dataset, mean across the ten  artificial datasets along with standard deviation in brackets is presented}
\label{tab:comp_defense}
\end{table*}

For each task, we will describe the defense mechanisms used, with the description of the replacement list, and replacement strategies. The calibration hyper-parameters used for the defense methods are temperatures $T=2$
%and 0.5\todo{change?} for scale 
across all tasks.
The canonical dictionary $M$ for NLI comprises of {\tt Microsoft}, {\tt Nasdaq} and {\tt IBM}.  
%
%\paragraph{NLI} For NLI, we use top frequent entities from the training set present in the training set for building our cannonicalization list $M$. The following names are used -- 
% \paragraph{GEC } 
For GEC, due to the huge size of the GEC corpus we pick the most common English first names and combine them with common English last names. We use three male names ({\tt  John,  James Brown, Robert Johnson}) and three female names ({\tt Patricia, Mary Jones, Jennifer Brown}) for replacement. If gender is ambiguous, we use 1 male name and 2 female names ({\tt John, Mary Jones, Jennifer Brown}). For CoRef, we used the top 3 frequent person names from the training dataset for our replacement list namely -- {\tt George Bush, Bill Clinton, Ehud Barak}. We also present results when we restrict the cannonical dictionary $M$ to only the first name in the above described lists.

%\todo[inline]{Did we use a single replacement  or all entries in $M$? --> all}

%For original dataset, nine random replacement datasets are created (using Algo. \ref{algo:attack}\todo{This makes no sense}) and mean and standard deviation across the ten datasets is provided.

We show results with RESEMBLE in Table~\ref{tab:comp_defense}. We perform defense on four datasets -- \texttt{Original, Best, Worst, Random Replacement}.  For random replacement, we constructed 10 new datasets from the original dataset with its names replaced with randomly selected names, and then evaluate the performance of our models on these datasets. We present the mean and standard deviation of the F scores across these newly constructed datasets. For best and worst we evaluate performance on datasets generated from Algo. \ref{algo:attack}. First observe that accuracy of even the original test dataset improves with our simple replacement ensembling while reducing the variance. For example, for GEC $F$ score increases from 50.93 to 51.81. The variance has also reduced as seen for the random replacement datasets.  The adversarial accuracy improves significantly --- for CoRef we see a jump of $D_\text{Worst}$ from 60.91 to 68.31 and for GEC the gains are even higher.  The difference between the best and worst accuracy reduces drastically. Although for $D_\text{Best}$ accuracy drops with RESEMBLE, the overall gains across the three dataset variants are much higher. Further a single canonical entry $M=1$ is almost as effective as larger ensembles of $M=3$. This implies that at test-time, we have to deploy the model on at most two instances to enjoy significantly higher robustness.
This shows that replacement with canonical entities while accounting for uncertainty of entity identification is a viable alternative to enhance robustness.
%not only can we recover almost all the drop in performance suffered from adversarial attacks but the accuracy on the original test set is also improved with our approach. Both the stability and variance of model is also reduced. One side effect of our approach is performance loss on the best dataset.
%\todo{Include stability numbers?}
%\begin{table*}[!h]
%\centering
%\begin{tabular}{|c|c|c|c|c|c|c|}
%\hline
%& \multicolumn{3}{c}{Cannonicalization} & \multicolumn{3}{c}{Graphical} \\ \hline
% & Nominal & Worst & Best  & Nominal & Worst & Best \\ \hline
%Google & 84.87 & 84.01 & 85.18 & 84.44 & 84.01 & 85.5 \\ \hline
%Facebook & \textbf{86.02} & \textbf{85.18} & 86.34 & \textbf{86.02} & 84.75 & 86.7  \\ \hline
%Microsoft & 85.81 & 84.98 & 86.13 & 85.81 & 84.98 & \textbf{86.86}  \\ \hline
%\end{tabular}
%\caption{Comparing Defense methods for NLI}
%\label{fig:nli_result}
%\end{table*}
%
%
%\subsubsection{CoRef}
%
%\begin{table*}[!h]
%\centering
%\begin{tabular}{|c|c|c|c|c|c|c|}
%\hline
%& \multicolumn{3}{c}{ Canonicalization } & \multicolumn{3}{c}{ Graphical } \\ \hline
%Name & Original & Worst & Best & Original & Worst & Best \\ \hline
%George Bush  & 77.05 & 76.88 & 76.24 & 78.54 & 78.15 & 78.17 \\ \hline
%Bill Clinton  & 77.71 & 77.78 & 78.01 & 78.28 & 78.11 & 77.88 \\ \hline
%Ehud Barak & 77.52 & 77.55 & 77.38 & 77.95 & 77.46 & 77.92 \\ \hline
%Slobodan Milosevic & 77.22 & 77.55 & 77.93 & 78.12 & 77.66 & 78.03 \\ \hline
%Vladimir Putin & 78.94 & 78.85 & 78.82 & 78.93 & 78.24 & 78.56 \\ \hline
%\end{tabular}
%
%\n{GRAPHICAL MODEL CoRef RESULTS}
%\label{fig:span_CoRef_result}
%\end{table*}

\section{Related Work}
%\todo[inline]{Let's move this to just before expts.}
% \subsection{Representing named entities}
% %\todo{Not looking good, drop this section?}
% Word embeddings can be categorized into various types of hierarchies such as non-contextual \cite{mikolov13, pennington2014glove, bojanowski2017fasttext} vs contextual embeddings \cite{peters2018deep, devlin2018bert} or word-level (word2vec,glove) vs character-level (ELMO) vs wordpiece-level embeddings (BERT). Different approaches described above will handle named entities in different fashion. For example, if word level embeddings will give 'UNK' to an OOV name while character level embeddings and wordpeice level embeddings will try to build a representation from constituting characters or wordpeices. Similarly, contextual models would exhibit different output for named entities across different sentences but at the same time would incorporate context from sentence and likely form richer name representations while non-contextual named entity would be static with respect sentences.

% In this work we try to see how well do the much appraised BERT entity representations fare which use wordpeice embeddings and contextual embeddings.

%  % Unlike adversarial attacks in computer vision or speech processing, care must be taken to ensure that the adversarial sentence retains its fluency. 
%

\paragraph{Study of BERT Representations}
\citet{jin2019bert, Lun2019onthe} study robustness of state of the art BERT fine-tuned models on classification, entailment, and machine translation tasks with respect to synonym replacements. The former used a black box scenario while the latter used input gradients and attention magnitudes to find probable candidate replacements. \citet{sun2020advbert} applied an adversarial mis-spelling attack to BERT using gradient-based saliencies. \citet{poerner2019} show that BERT uses the surface form of words for relational reasoning (guessing person with an Italian sounding name speaks Italian). \citet{zhang2019paws} generated adversarial sentence pairs for paraphrase detection by swapping the order of named entities in two sentences which was enough to fool BERT. \citet{joshi2019spanbert} introduced Span-BERT that is trained on masked language modelling on spans instead of tokens. \citet{zhang2019ernie} developed ERNIE model for entity linking which combines named entity embeddings from knowledge graph with BERT. % %\todo[inline]{Give more details of these methods.  Add more papers that study various aspects of BERT representations. Include Span-BERT, ERNIE and other entity focused BERT variants.} 

\paragraph{Other Robustness Studies in NLP}
Techniques for generating adversarial examples to study robustness of NLP models have seen a lot of enthusiasm in recent years. These approaches can be loosely categorized into three types -- character-level~\cite{hotflip,hotflipnmt} or word-level or sentence-level~\cite{zhao2018generating,iyyer2018paraphrase,ribeiro2018}. Our work is most related to word-level attacks which we elaborate on.
%\paragraph{Character Level Attacks} - \citet{hotflip,hotflipnmt} study gradient-based string edits to fool classification and translation systems respectively. 
%
%\paragraph{Word Level Attacks} - 
\citet{liangfooled18} proposed word insertion, deletion, or replacement using gradient magnitudes for classification tasks but requires human effort to ensure the sensibility of the replacements. \citet{samandmehta17} used synonym replacements along with the gradient sign method for choosing the worst synonym replacement. \citet{alzantot2018} provides a population-based genetic algorithm for synonym attacks for sentiment classification and textual entailment in a black-box setting. \citet{PWWS} developed a greedy algorithm for synonym swaps using weighted gradient based word saliencies, for sentiment classification and entailment. 

%\paragraph{Sentence Level Attacks} - \citet{zhao2018generating} used a generative adversarial network (GAN) for creating "natural" adversarial examples. \citet{iyyer2018paraphrase} used syntactically controlled paraphrase networks (SCPNs) which break pretrained models. \citet{ribeiro2018} proposed SEARS, phrase-level adversarial sentence transformation rules for debugging current NLP systems. Recently for paraphrase detection 
%
In this work, we also perform word-level attacks but our focus is robustness to named entity replacements.  The closest work to ours is \cite{sensitivitybiases} that checks the sensitivity of models with respect to named entities but they only consider sentiment or toxicity classification.  Our work covers more interesting structured prediction tasks such as coreference resolution and grammatical error correction. 

\paragraph{Defenses in NLP}
Most approaches \cite{cheng2018towards,jia2017adversarial} for defenses in NLP have focused on augmenting training datasets with adversarial instances. \citet{pruthi2019combating} proposed a word recognition model along with backoff strategies for robustness against misspellings. \citet{zhou2019learning} used an adversarial detection cum replacement strategy.  We did not consider data augmentation methods because that would significantly increase the training time for models like GEC.  

There has also been a trend in usage of certified robustness approaches \cite{yun2019popqorn,jia2019certified,huang2019achieving, Shi2020Robustness} which provide guarantees on the minimum performance of models.  The main technique so far is to propagate interval bounds around input word embeddings and has been applied for robustness to synonyms change. Synonyms are expected to have similar embeddings, but interval bounds are unlikely to work for entities within a large concept class.
We are not aware of any prior work that enhances robustness with canonical replacements like ours in the context of an existing language model.

\section{Conclusions and Future Work}
We show that state of the art BERT-based models are surprisingly brittle to named entity replacements. RESEMBLE, a simple ensembling approach increases robustness while also improving the nominal accuracy. The general paradigm of enhancing robustness via ensembles on guided instance perturbations is a promising direction and needs to be explored for other tasks too.

\paragraph{Acknowledgement}
We thank the anonymous reviewers for their constructive feedback on this work. This research was partly sponsored by IBM AI Horizon Networks - IIT Bombay initiative and partly by a Google India AI/ML Research Award. Abhijeet is supported by Google PhD Fellowship %in Machine Learning.

\bibliographystyle{acl_natbib}
\bibliography{ML}

\newpage

\begin{appendices}

\section{}%

\subsection{Universal named entity attacks}

We then tried to investigate if universally adversarial names exist, that is, names that induce an incorrect output from the model in most inputs. 

We modeled this as a search problem, given a large dataset ($D$) and a large list of names ($V$) we try to find a name with the worst performance and claim it to be the adversarial name. Since $V\times D$ could be very large we resort to sampling methods to reduce the search space. Our algorithm can be found in \ref{algo:univ}. We keep a set of parameters for each name which models the prior distribution of the fall in score when the names are replaced by the given name. The Evaluate function computes the performance of the model given a dataset and outputs per sentence score. Then using these scores we update the per-name parameters in accordance with Bayes' theorem and recompute and update weights according to the new parameters.

\begin{algorithm}[!h]
\SetAlgoNoLine
\SetArgSty{textnormal}
\newcommand\mycommfont[1]{\footnotesize\ttfamily\textcolor{blue}{#1}}
\SetCommentSty{mycommfont}
\KwData{$V$(superset of names), $D$(dataset to evaluate on) , $M$(model), $B$(budget), $N$(iterations), $ip$(initial prior parameters)}
\KwResult{$W$ (weights for each name correlating with "adversarial"-ness of the name)}

$L = len\left( V \right)$ ;\\
$W \gets \text{array}\left( value= \frac{1}{L}, length=L \right)$;\\
$P \gets \text{array}\left( value= ip, length=L \right)$;\\
$map \gets \text{dict}\left( keys=V, values=[1..L] \right)$;\\
\For{$i = 0$ to $N$ }{ 
    $V_s \gets \text{RandomSample}(V, weights=W, sample\_size=B)$;\\
    $D_s \gets \text{RandomSample}(D, weights=W, sample\_size=B)$;\\
    $osc \gets \text{Evaluate}(M, D_{s} )$;\\
   \For{$v \in V_s$}{
        $D_{s}^{new} \gets \text{Replace}(D_s, v)$;\\
        $sc \gets \text{Evaluate}(M, D_{s}^{new} )$;\\
        $\text{UpdateWeightsParameters}(W, P,osc, sc,$\\  $index=map[v])$;\\
   }
}

\caption{Finding universal named entity triggers}
\label{algo:univ}
\end{algorithm}

We apply our algorithm on the GEC tasks since it exhibited a considerable amount of non-robustness and low stability. To start we used about 5,000 first names as $V$. We use the locness train dataset to find top candidates for adversarial names and also see how such names generalize to other datasets (here locness test). We model the weights as the probability that a particular name would decrease the score minus that a name increases the score. We parameterize this probability with a beta distribution. We initialize $\alpha = \beta = 1$ and at every step of algorithm, update these parameters. So if $20\%$ of the $B$ sentences in the sampled dataset get their predictions worsened and $10\%$ of $B$ sentences predictions improved, then we update $\alpha \longleftarrow \alpha + 0.1B$, and $\beta \longleftarrow \beta + 0.9B$. We did not find any existence of universal named entity triggers for GEC. We trained weights using the above algorithm and use the top 10 candidate names from our algorithm. We found that the performance dropped by only about 1-1.5 points. Then we also tried to see the generalization of these names on the test-common dataset and found that these candidate names do not generalize and performance drops by only 0.1/0.2, which is staying virtually the same!
,

\end{appendices}

\end{document}